\documentclass[a4paper, 10pt, conference]{ieeeconf}      
\usepackage{FG2020}

\FGfinalcopy 

\usepackage{epsfig} 
\usepackage{amsmath} 
\usepackage{amssymb}  
\usepackage{array}
\usepackage{booktabs}
\usepackage{multirow}
\usepackage[noadjust]{cite}
\usepackage{pgfplots}
\usepgfplotslibrary{groupplots}
\newcolumntype{L}{>{$}l<{$}}
\newcolumntype{C}{>{$}c<{$}}
\newcolumntype{R}{>{$}r<{$}}

\DeclareMathOperator{\sigmoid}{sigmoid}

\def\FGPaperID{2 (RFIW)} 

\title{\LARGE \bf
A Multi-Task Comparator Framework for Kinship Verification}

\author{\parbox{16cm}{\centering
    {\large Stefan H\"ormann, Martin Knoche, Gerhard Rigoll}\\
    {\normalsize
    Chair of Human-Machine Communication, Technical University of Munich, Germany\\}}
}

\begin{document}

\ifFGfinal
\thispagestyle{empty}
\pagestyle{empty}
\else
\author{Anonymous FG2020 submission\\ Paper ID \FGPaperID \\}
\pagestyle{plain}
\fi
\maketitle

\begin{abstract}
Approaches for kinship verification often rely on cosine distances between face identification features. However, due to gender bias inherent in these features, it is hard to reliably predict whether two opposite-gender pairs are related. Instead of fine tuning the feature extractor network on kinship verification, we propose a comparator network to cope with this bias. After concatenating both features, cascaded local expert networks extract the information most relevant for their corresponding kinship relation. We demonstrate that our framework is robust against this gender bias and achieves comparable results on two tracks of the RFIW Challenge 2020. Moreover, we show how our framework can be further extended to handle partially known or unknown kinship relations.
\end{abstract}

\vspace{0.08cm}
\section{INTRODUCTION}
Kinship relationship between two people is usually determined using the persons' physical features, which can be divided into DNA, body and facial features. In contrast to the very reliable DNA-analysis, facial- and body features are used to obtain an initial and quick estimate of whether two people are related or not.

Image-based kinship verification \cite{georgopoulos2018modeling,fang2010towards,guo2012kinship,yan2014discriminative,yan2014prototype,xia2012understanding,dibeklioglu2013like,wang2014leveraging,dawson2018same,wang2018cross,robinson2018visual,rachmadi2018paralel,rachmadi2019image,dahan2017kin,wang2017kinship,robinson2016fiw,laiadi2019kinship,hu2014large,xia2012toward,zhou2011kinship,zhou2012gabor,patil2019deep,robinson2017recognizing} relies only on information present in facial images to estimate whether they are related. Due to the inherent flexibility of only needing a face image compared to more invasive DNA-sample, kinship verification with visual media has an abundance of practical uses: e.g., forensic investigations, genealogical studies, social media-based analysis and photo library management. As proposed in the RFIW Challenge 2020 \cite{robinson2020recognizing}, one can state three problems concerning kinship verification:
\begin{enumerate}
    \item Determine whether two persons are consanguine given a kinship relation.
    \item Decide whether a person is the child of given parents.
    \item Identifying relatives of a person in a gallery.
\end{enumerate}

Lately, the emergence of bigger image kinship verification datasets, including CornellKinFace \cite{fang2010towards},  KinFaceW \cite{lu2013neighborhood,lu2015fg}, TSKinFace \cite{qin2015tri}, and FIW\cite{robinson2016fiw}, has given more and more attention to kinship-related tasks and allowed the development of more reliable data-based approaches. Kinship verification from face images focuses on consanguinity kinship, which can be divided into three groups:
\begin{itemize}
    \item Same-generation pairs: brother-brother \textit{BB}, brother-sister \textit{SIBS} and sister-sister \textit{SS}
    \item First-generation pairs: father-son \textit{FS}, father-daughter \textit{FD}, mother-son \textit{MS} and mother-daughter \textit{MD}
    \item Second-generation pairs:  grandfather-grandson \textit{GFGS}, grandfather-granddaughter \textit{GFGD}, grandmother-grandson \textit{GMGS} and grandmother-granddaughter \textit{GMGD}
\end{itemize}

\begin{figure}[t]
\centering
\vspace{0.25cm}
\includegraphics[width=0.95\columnwidth]{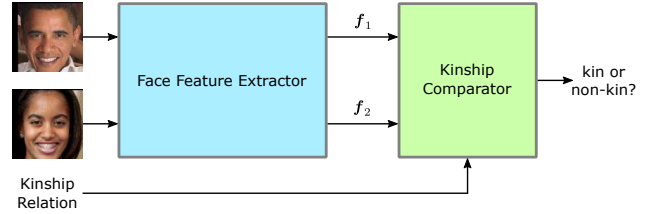}
\caption{Overview of the kinship recognition comparator framework: Features $\boldsymbol{f}_{\text{1}}$, $\boldsymbol{f}_{\text{2}}$ are extracted from two input faces, which are then combined in the comparator network to estimate whether the faces are related according to a given kinship relation.}
\vspace{-0.3cm}
\label{fig:network_small}
\end{figure} 
\begin{figure}[t]
\newlength\figureheight
\newlength\figurewidth
\setlength\figureheight{4.5cm}
\setlength\figurewidth{0.59\columnwidth}

\include{figures/histo_60}
\vspace{-0.6cm}
\caption{Histogram of the cosine distance of face identification features $\boldsymbol{f}_{\text{1}}$ and $\boldsymbol{f}_{\text{2}}$ for \textit{kin} (blue) and \textit{non-kin} (orange) pairs for parents-daughter kinship relations on the RFIW validation dataset. Best viewed in color.}
\vspace{-0.5cm}
\label{fig:histogram}
\end{figure}

\begin{figure*}[t]
\centering
\includegraphics[width=0.94\textwidth]{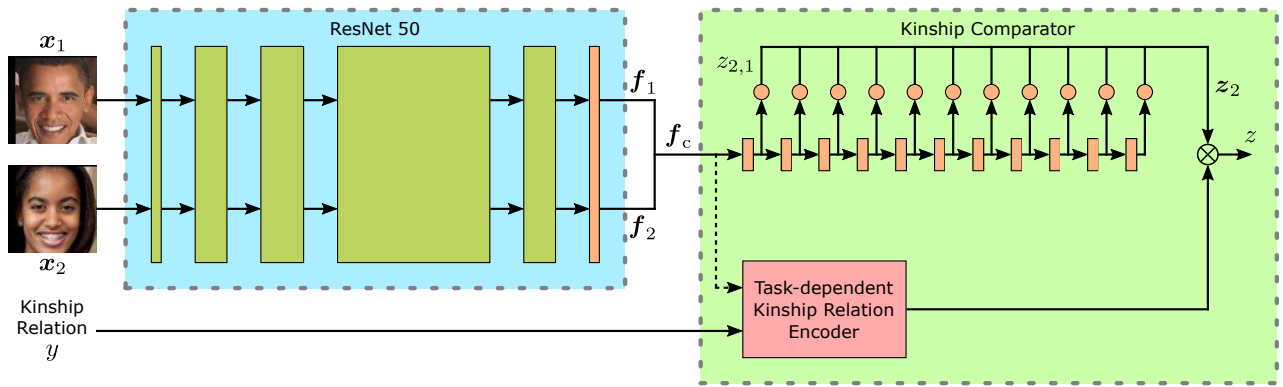}
\caption{Our multi-task kinship comparator framework: Faces are embedded separately into a feature space using a ResNet-50 (ResNet blocks in green). The concatenation of both output features $\boldsymbol{f}_{\text{c}}$ is used by a cascaded local expert network, which uses multiple fully connected layers (orange) to refine the information and focus it into a single neuron (orange circle) for each kinship relation. The task-dependent kinship relation encoder selects the output neuron corresponding to the given kinship layer, which will be forwarded to the output $z$. Best viewed in color.}
\vspace{-0.3cm}
\label{fig:network}
\end{figure*}

As illustrated in Fig. \ref{fig:network_small}, typical kinship verification approaches consist of a convolutional neural network, which extracts facial features for each image separately. These features are then fed into a kinship comparator in order to distinguish between \textit{kin} or \textit{non-kin}. Several methods \cite{robinson2018visual,laiadi2019kinship, robinson2020recognizing} rely on metrics like cosine distance between extracted features to determine kinship. However, as shown in the histogram in Fig.\ref{fig:histogram}, \textit{kin} and \textit{non-kin} pairs from opposite-gender kinship relations are hardly separable compared to same-gender kinship relations, which is due to the high influence of gender on the feature. 

Motivated by this finding and in contrast to training the feature extractor on kinship recognition \cite{robinson2018visual,dahan2017kin,rachmadi2019image2,aspandi2019heatmap,patil2019deep}, we propose a comparator framework, which is robust against this gender bias as we demonstrate later. Our neural kinship comparator framework is not only capable of solving typical kinship-related tasks benefiting from separated local expert networks for each kinship relation but can also be further extended with an attention module to predict the kinship relation and leverage it for tasks with unknown kinship relation.


\section{RELATED WORK}
According to Georgopoulos et. al. \cite{georgopoulos2018modeling} approaches for kinship verification can be divided into five categories:

Invariant descriptors based methods \cite{guo2012kinship, fang2010towards} are focusing on how to represent local facial parts. Subspace learning-based approaches \cite{yan2014discriminative,yan2014prototype} learn a kinship invariant subspace capitalizing on techniques like factor analysis and transfer learning. Metric learning-based methods \cite{wang2009information,davis2007information,ding2016robust} involve learning a distance measure or feature transformation, and are used to reduce the feature distance between \textit{kin} pairs while extended the distance for \textit{non-kin} pairs. Approaches using contextual and dynamic features \cite{xia2012understanding,dibeklioglu2013like}, applying texture descriptors and using geometric information \cite{wang2014leveraging} have also been studied for kinship verification.

Apart from these traditional methods, deep learning-based  approaches lately achieved state-of-the-art performance in kinship verification. Widely used architectures, e.g., VGG16 \cite{dawson2018same}, GAN \& ResNet \cite{wang2018cross}, SphereFace \cite{robinson2018visual}, SPCNN \cite{rachmadi2018paralel}, ShallowResNet \cite{rachmadi2019image} and VGGFace \cite{dahan2017kin,wang2017kinship,robinson2016fiw} have been used for this task. Laiadi et. al \cite{laiadi2019kinship} proposed a novel approach feeding the cosine similarity, which is computed from deep (VGG-Face descriptor) and tensor (BSIF- and LPQ-tensor using MSIDA method) features,  through an extreme learning machine in order to verify kinship. 

\begin{table*}[t]
  \centering
  \caption{Accuracy of the top 10 results on the RFIW kinship verification challenge dataset separated by kinship relation $y$. The best result for every kinship relation is marked with $*$ and bold with the runner-up being denoted by $\diamond$.}
    \begin{tabular}{lLLLLLLLLLLLL}
    \toprule
    & \multicolumn{11}{c}{Accuracy~[\%]}\\
    \cmidrule(lr){2-13}
    & & \multicolumn{3}{c}{siblings} & \multicolumn{4}{c}{parent-child} & \multicolumn{4}{c}{grandparent-grandchild} \\
    \cmidrule(lr){3-5} \cmidrule(lr){6-9} \cmidrule(lr){10-13} 
    User  & \text{Average} & \textit{BB} & \textit{SIBS} & \textit{SS} & \textit{FD} & \textit{FS} & \textit{MD} & \textit{MS} & \textit{GFGD} & \textit{GFGS} & \textit{GMGD} & \textit{GMGS}   \\
          \midrule
    vuvko & \textbf{78.1}~* & \textbf{80.2}~* & \textbf{77.3}~* & \textbf{80.4}~* & 75.2 & 80.8 & \textbf{77.7}~* & 74.4 & 77.9~\diamond & 69.4 & \textbf{75.8}~* & 59.8 \\
    DeepBlueAI & 76.1~\diamond & 76.5~\diamond & 74.6 & 76.9~\diamond & 74.4 & 80.8 & 75.1 & 73.9 & 72.5 & 72.7 & 67.3 & \textbf{67.6}~* \\
    ustc-nelslip & 75.9 & 75.1 & 72.0 & 74.4 & 75.5~\diamond & \textbf{81.8}~* & 74.7 & 75.2 & \textbf{78.6}~* & 69.0 & \textbf{75.8}~* & 67.0~\diamond \\
    haoxl & 75.5 & 74.8 & 71.1 & 74.0 & 75.5~\diamond & 81.2 & 74.7 & 75.2 & 72.9 & 64.9 & 63.2 & 64.3 \\
    lemoner20 & 75.4 & 75.0 & 72.2 & 74.5 & 75.4 & 80.7 & 74.0 & 75.0 & 72.0 & 66.9 & 61.7 & 65.4 \\
    \midrule
    Early & 74.2 & 74.6 & 72.9 & 74.3 & 73.4 & 78.5 & 72.3 & 74.4 & 65.7 & 68.6 & 52.4 & 64.8 \\
    \textbf{ours} & 73.6 & 66.4 & 76.0~\diamond & 65.3 & \textbf{76.9}~* & 80.1 & 76.7~\diamond & \textbf{78.2}~* & 70.0 & 73.4~\diamond & 63.9 & 60.3 \\
    bestone & 73.2 & 69.2 & 62.4 & 67.1 & 75.4 & 81.2~\diamond & 75.4 & 75.4~\diamond & 73.1 & 69.4 & 64.7 & 62.0 \\
    danbo3004 & 72.6 & 71.3 & 70.9 & 72.0 & 72.4 & 78.1 & 71.5 & 72.0 & 71.1 & 69.8 & 53.2 & 56.4 \\
    ten\textunderscore elven & 72.3 & 72.2 & 71.4 & 73.4 & 70.1 & 77.1 & 70.0 & 71.2 & 69.8 & \textbf{74.7}~* & 63.2 & 67.0~\diamond \\
    \bottomrule
    \end{tabular}%
  \label{tab:track1}%
  \vspace{-0.3cm}
\end{table*}%
\section{MULTI-TASK KINSHIP COMPARATOR FRAMEWORK}

\subsection{Face Feature Extractor}
\label{sec:faceextractor}

We embed face images into a deep feature space using the adapted ResNet-50 \cite{he2016identity} with the ArcFace layer according to \cite{deng2019arcface}. To pretrain the model with softmax cross-entropy on the refined MS-Celeb-1M dataset \cite{MS1M}  we add an 85164-dimensional fully connected layer, which is dropped later on together with the ArcFace layer. This bottleneck architecture together with the ArcFace layer ensures a well-generalizing identity feature vector $\boldsymbol{f} \in \mathbb{R}^{512}$.

\subsection{Kinship Comparator}
\label{sec:track1}
Generally, we can describe the kinship verification task as follows: Given the triplets  $\left(\boldsymbol{x}_1,\,\boldsymbol{x}_2, \,y\right)$ consisting of two images $\boldsymbol{x}_1,\,\boldsymbol{x}_2 \in \mathbb{R}^{112\times112\times3}$ and a kinship relation $y$, the goal is to determine whether $\boldsymbol{x}_1$ and $\boldsymbol{x}_2$ are related as encoded in $y$, which is denoted by the probability $z$ at the output of our framework. Both face images are embedded independently by the feature extractor described in subsection \ref{sec:faceextractor} yielding the corresponding feature vectors $\boldsymbol{f}_1$ and $\boldsymbol{f}_2$.

For our kinship comparator, depicted in Fig. \ref{fig:network}, we concatenate both features $\boldsymbol{f}_{\text{c}} = \left[\boldsymbol{f}_1,\, \boldsymbol{f}_2\right]$. Next, the concatenated feature vector is fed into the first out of eleven local expert networks. By building a local expert for every kinship relation $y$ we allow every local expert to focus only on parts of the features relevant of its corresponding kinship relation. For same-gender kinship relations (\textit{BB}, \textit{SS}, \textit{FS}, \textit{MD}, \textit{GFGS} and \textit{GMGD}) the local expert can deduce from detecting separate genders that both input images cannot be related. Similarily, by detecting same gender a opposite-gender kinship relation (\textit{SIBS}, \textit{FD}, \textit{MS}, \textit{GFGD} and \textit{GMGS}) can be excluded. Moreover, different facial features are shared from mother or father of a child affirming our proposed architecture.

Every local expert is an identical fully connected neural network with two layers. The first layer consists of 192 neurons with leaky ReLU \cite{maas2013rectifier} as activation function. We conclude the local expert network with a fully connected layer consisting of a single neuron and sigmoid activation function to obtain a probability $z_{2,i}$  for the $i$-th local expert between 0 and 1. While the input of the first local expert is the concatenated feature vector $\boldsymbol{f}_{\text{c}}$, the remaining local experts use the output of the previous local experts $z_{1,i-1}$. Due to this architecture, the information is first refined in every 192-dimensional layer and then the information most relevant for its specific kinship relation is extracted in the second 1-dimensional layer.
Mathematically, the output of the $i$-th local expert $z_{2,i}$ can be formulated as follows:
\begin{align}
    &\boldsymbol{z}_{1,1} = \max \left(\boldsymbol{W}_1 \boldsymbol{f}_{\text{c}} + \boldsymbol{b}_1, 0.2 \cdot \left(\boldsymbol{W}_1 \boldsymbol{f}_{\text{c}} + \boldsymbol{b}_1\right)\right)\\
    &\boldsymbol{z}_{1,i} = \max \left(\boldsymbol{W}_1 \boldsymbol{z}_{1,i-1} + \boldsymbol{b}_1, 0.2 \cdot \left(\boldsymbol{W}_1 \boldsymbol{z}_{1,i-1} + \boldsymbol{b}_1\right)\right) \label{eq:2}\\
    &z_{2,i} = \sigmoid\left(\boldsymbol{W}_2 \boldsymbol{z}_{1,i} + b_2\right)
\end{align}
with $\boldsymbol{W}_1$, $\boldsymbol{W}_2$ and $\boldsymbol{b}_1$, $b_2$ denoting the trainable weight matrix and bias (vector) of the first and second layer, respectively. Note that, (\ref{eq:2}) is only valid for $i > 1$.

By concatenating the outputs of all local experts $\boldsymbol{z}_2 = \left[z_{2, 1},\,\cdots,\, z_{2,11}\right]$ we obtain a probability of $\boldsymbol{x}_1$ and  $\boldsymbol{x}_2$ being related for every kinship relation $y$. Since the kinship relation $y$ is given for the kinship verification task, we can observe its corresponding probability at the output $z$ of the framework by performing a scalar multiplication of $\boldsymbol{z}_2$ with the one-hot encoding of $y$. We can also interpret this one-hot encoding, which is generated by the task-dependent kinship relation encoder depicted in Fig. \ref{fig:network}, as relying entirely on the output of the local expert selected by $y$. Later, we will show that this disentanglement between the predictors $z_{2,i}$ and the selection of the predictors according to an ideally given kinship relation offers a variety of opportunities for future extensions. Another benefit of this structure is the joint training of all local experts without restricting the capabilities of the framework performing multiple tasks.

\subsection{Extension for Tri-Subject Verification}
\label{sec:track2}

For the tri-subject verification task, a quadruple $\left(\boldsymbol{x}_1,\,\boldsymbol{x}_2,\,\boldsymbol{x}_3, \,y'\right)$ is given with $\boldsymbol{x}_1$, $\boldsymbol{x}_2$ and $\boldsymbol{x}_3$ indicating the image of the father, mother and child, respectively, and $y'$ denoting the gender of the child. In accordance with subsection \ref{sec:track1}, this task can be performed by our framework by splitting the sample into two separate triplets $\left(\boldsymbol{x}_1,\,\boldsymbol{x}_3, \,\textit{FC}\right)$ and $\left(\boldsymbol{x}_2,\,\boldsymbol{x}_3, \,\textit{MC}\right)$ with \textit{C} being a placeholder for whether the child is the parents' son \textit{S} or daughter \textit{D}. Feeding these triplets into our framework, we obtain two separate probabilities $z_{\textit{FC}}$ and $z_{\textit{MC}}$ indicating how likely it is that the child is related to the father and mother, respectively. 

\section{EXPERIMENTS}
\subsection{Training Details}
\label{sec:train}

Training is divided into two stages:
We pretrain the face feature extractor with softmax cross-entropy loss for face identification using the MS-Celeb-1M dataset, which contains over $5.8$\,M images of over 85k identities \cite{MS1M}. As preprocessing, we align all faces with facial landmarks predicted by the MTCNN \cite{zhang2016MTCNN} and crop them afterwards to $112 \times 112$ pixels. Our face feature extractor achieves an accuracy of $99.63\,\%$ on the LFW benchmark \cite{LFW}.

While keeping the weights of the face feature extractor constant, we train the kinship comparator using pairs generated from the RFIW training dataset \cite{robinson2020recognizing} consisting of $\approx 249$\,k \textit{kin} pairs (after duplicating and swapping $\boldsymbol{x}_2$ with $\boldsymbol{x}_2$ for all same-generation pairs  \textit{BB}, \textit{SS} and \textit{SIBS}). In order to generate meaningful \textit{non-kin} pairs, we randomly swap $\boldsymbol{x}_2$ with $\boldsymbol{x}_2$ from a different family with the same kinship label $y$ in every epoch. By doing so, we ensure not only high variety among the \textit{non-kin} pairs but also that gender and age of the \textit{non-kin} pairs match the kinship relation $y$. We preprocess the faces identically as when pretraining the face feature extractor. As data augmentation, we perform left-right flipping, and random contrast, brightness and saturation with a probability of $50\,\%$. The kinship comparator is trained on binary sigmoid cross-entropy loss for 4 epochs using the ADAM-optimizer \cite{ADAM} with a batch size of 200 and an initial learning rate of $0.001$, which we decrease to $0.0005$ after the second epoch. In order to improve generalization we add $20\,\%$ dropout on the concatenated feature vector $\boldsymbol{f}_{\text{c}}$ and train with an additional regularization loss on the $L^2$-Norm of all trainable weights of the kinship comparator with a factor of $2 \cdot 10^{-4}$.

\subsection{Results: RFIW Track 1 - Kinship Verification}
We report our performance evaluating on the RFIW challenge dataset \cite{robinson2020recognizing} containing $\approx 40$\,k image pairs with their kinship labels. In order to decide whether the probability $z$ is sufficient to classify a pair as related, we compute the threshold which yields the best average accuracy on the validation set, which consists of $\approx 129$\,k pairs in total  (after creating \textit{non-kin} pairs as described in subsection \ref{sec:train}). 

\begin{figure}[t]
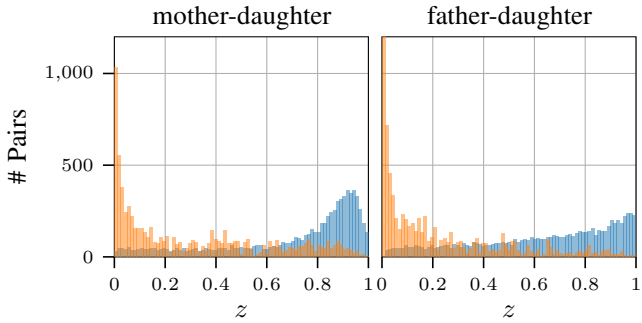

\setlength\figureheight{4.5cm}
\setlength\figurewidth{0.57\columnwidth}

\include{figures/logit_70}
\vspace{-0.6cm}
\caption{Histogram of the output $z$ of \textit{kin} (blue) and \textit{non-kin} (orange) pairs for parents-daughter kinship relations on the RFIW validation dataset. Best viewed in color.}
\label{fig:histogram_logit}
\end{figure}

Table \ref{tab:track1} illustrates the accuracy for each kinship relation $y$. Even though our approach does not yield the best average accuracy, our average accuracy is only $2.5$\,\% lower than second place. Moreover, we outperform all other approaches for the kinship relation father-daughter \textit{FD} and mother-son \textit{MS}, and further obtain second-best accuracy on three more kinship relations (\textit{SIBS}, \textit{MD} and \textit{GFGS}). This indicates that our approach stands out especially in opposite-gender kinship relations (\textit{FD}, \textit{MS} and \textit{SIBS}), which is also affirmed by the better separable distributions of \textit{kin} and \textit{non-kin} pairs using the output $z$ as shown in Fig. \ref{fig:histogram_logit} compared to the cosine distance in Fig.  \ref{fig:histogram}. The inferior performance on other opposite-gender kinship relations (\textit{GFGD} and \textit{GMGS}) can be explained by the substantially smaller amount of training data ($\approx 4$\,k pairs for grandparent-grandchild relations compared to $61\,\text{k} -  94$\,k pairs for \textit{SIBS}, \textit{FD} and \textit{MS}).

Table \ref{tab:track1_param} shows the accuracy of our framework for different activation functions, dropout probabilities and hidden layer sizes on the RFIW kinship verification validation dataset with the \textit{non-kin} pairs generated as mentioned in subsection \ref{sec:train}. It can be seen that a dropout of 20\,\% - 40\,\% together with leaky ReLU as activation function yields the best results. However, a higher amount of neurons in the hidden layer seems very likely to boost the performance on the challenge dataset even further.

\begin{table}[t]
  \centering
  \caption{Average accuracy on RFIW kinship verification validation dataset for different parameter settings with the last row being the same model as in Table \ref{tab:track1}.}
    \begin{tabular}{cCCC}
    \toprule
    Activation Function & \text{Dropout} & \text{Layer Size} & \text{Accuracy~[\%]} \\
          \midrule
    ReLU  & 20\,\%  & 192   & 76.8 \\
    PReLU & 20\,\%  & 192   & 75.6 \\
    Tanh  & 20\,\%  & 192   & 72.5 \\
          \midrule
    LReLU & \phantom{2}0\,\%   & 192   & 75.5 \\
    LReLU & 10\,\%  & 192   & 76.6 \\
    LReLU & 30\,\%  & 192   & 77.7 \\
    LReLU & 40\,\%  & 192   & 77.1 \\
          \midrule
    LReLU & 20\,\%  & 64    & 74.5 \\
    LReLU & 20\,\%  & 128   & 77.5 \\
    LReLU & 20\,\%  & 256   & 77.2 \\
    LReLU & 20\,\%  & 512   &\textbf{79.6} \\
    LReLU & 20\,\%  & 1024  & 78.8 \\
          \midrule
          \midrule
    LReLU & 20\,\%  & 192   & 77.5 \\
    \bottomrule
    \end{tabular}%
    \vspace{-0.3cm}
  \label{tab:track1_param}%
\end{table}%

\subsection{Results: RFIW Track 2 - Tri-Subject Verification}

As mentioned in subsection \ref{sec:track2}, by splitting the tri-subject verifications into two verification problems we obtain two probabilities $z_{\textit{FC}}$ and $z_{\textit{MC}}$. Since according to Table \ref{tab:track1} there is only a small difference between same-gender and opposite-gender parent-child pairs, we take the average of both probabilities and use the threshold obtaining the best accuracy on the RFIW tri-subject validation set. The results on the RFIW tri-subject challenge dataset are depicted by Table \ref{tab:track2} showing that our method yields comparable results.

\begin{table}[h]
  \centering
  \caption{Accuracy on the RFIW tri-subject verification challenge dataset.}
    \begin{tabular}{lLLL}
    \toprule
    & \multicolumn{3}{c}{Accuracy}\\
    \cmidrule(lr){2-4}
    User  & \text{Average} & \text{FMD}   & \text{FMS} \\
          \midrule
    ustc-nelslip & 0.79  & 0.78  & 0.80 \\
    lemoner20 & 0.78  & 0.76  & 0.80 \\
    DeepBlueAI & 0.77  & 0.76  & 0.77 \\
    Early & 0.77  & 0.76  & 0.77 \\
    \textbf{ours} & 0.73  & 0.72  & 0.74 \\
    Ferryman & 0.72  & 0.70  & 0.74 \\
    will\textunderscore go & 0.68  & 0.66  & 0.70 \\
    \bottomrule
    \end{tabular}%
    \vspace{-0.3cm}
  \label{tab:track2}%
\end{table}%

\section{CONCLUSIONS AND FUTURE WORKS}

In this paper, we present a novel framework for multi-task kinship recognition, which achieves top accuracy for five out of eleven kinship relations compared to over 20 state-of-the-art methods on the RFIW kinship verification challenge dataset. The key advantage of our architecture is the joint training of a local expert network for each kinship relation. This not only allows every expert to extract the information necessary to reliably predict its corresponding kinship but also shares and refines information among the experts. Our approach performs especially well on opposite-gender pairs, which is affirmed by the reduction of gender bias originally present in face identification features. Moreover, we demonstrate that our framework achieves comparable performance on the tri-subject verification task.

The future work is twofold: First, as already indicated by the results Table \ref{tab:track1_param} a tuning of the local expert networks can further increase the performance. Entirely local experts with both fully connected layers being separated from each other have also shown their potential in our experiments, but tend to overfit due to the missing consecutive refinement in the shared first layer. For the tri-subject verification task, a more sophisticated pooling operation based on the confidence of each probability would be capable of fusing both probabilities more reliably.

Next, we plan to demonstrate the full potential of our approach by evaluating the performance additionally for partly known kinship relations. For instance, instead of using the kinship relation a more realistic scenario of knowing only the gender of both input images could be considered. Besides, the kinship relation can be unknown as in track 3 of the RFIW challenge. Even though first experiments have shown that average/max pooling of $\boldsymbol{z}_2$ does not yield satisfactory results, we propose to use the task-dependent kinship relation encoder as an attention module, which predicts the kinship relation based on the concatenated feature vector $\boldsymbol{f}_{\text{c}}$ (indicated by the dashed arrow in Fig. \ref{fig:network}). The first results indicate that the kinship relation can be correctly identified with an accuracy of at least $65\,\%$.

\addtolength{\textheight}{-1cm}   


\bibliographystyle{ieeetr}
\bibliography{ref}

\end{document}